\def\year{2020}\relax
\documentclass[letterpaper]{article} 
\usepackage{aaai20}  
\usepackage{times}  
\usepackage{helvet} 
\usepackage{courier}  
\usepackage[hyphens]{url}  
\usepackage{graphicx} 
\usepackage{graphicx}  
\usepackage{amsmath}
\usepackage{tikz}

\title{Improving the Performance of the LSTM and HMM Model via Hybridization}
\author{Larkin Liu\textsuperscript{\rm 1}, \Large \textbf{Yu-Chung Lin\textsuperscript{\rm 1},} \Large \textbf{Joshua Reid \textsuperscript{\rm 2}}\\ 
\textsuperscript{\rm 1}University of Toronto\\ 
\textsuperscript{\rm 2}University of Waterloo\\ 
\textsuperscript{\rm 1}\{larkin.liu, kohjerry.lin\}@mail.utoronto.ca, \textsuperscript{\rm 2}js2reid@uwaterloo.ca  
}
 
\begin{document}

\maketitle

\begin{abstract}
Language models based on deep neural networks and traditional stochastic modelling have become both highly functional and effective in recent times. In this work, a general survey into the two types of language modelling is conducted. We investigate the effectiveness of the Hidden Markov Model (HMM), and the Long Short-Term Memory Model (LSTM). We analyze the hidden state structures common to both models, and present an analysis on structural similarity of the hidden states, common to both HMM's and LSTM's. We compare the LSTM's predictive accuracy and hidden state output with respect to the HMM for a varying number of hidden states. In this work, we justify that the less complex HMM can serve as an appropriate approximation of the LSTM model.
\end{abstract}

\section{Introduction}

Language modelling has been an integral part of providing an understanding of the nature of language sequences. In order to improve the machine understanding of language using sequential models, we seek to explore two prominent areas of statistical language models, the Hidden Markov Model (HMM), and a Recurrent Neural Network (RNN) architecture, known commonly as Long Short-Term Memory (LSTM) Model. Under a discrete stochastic modelling framework, HMM's were first introduced in \cite{Rabiner:1987} to classify speech signals. First used to automate AT\&T's voice activated call center, the revolutionary technology allowed computers to robustly characterise speech, and form a basic understanding of spoken words. HMM's have since become a definitive benchmark for the state-of-the-art for speech recognition, and text recognition algorithms. Around the same period, RNN's were introduced by \cite{Rumelhart:1988}, however, the training complexity of the model was far too high and was not commensurate with the hardware capabilities at the time. In the 21st century, with the introduction of the more advanced hardware for deep learning, came a wave of applications for the RNN for both voice, text recognition  \cite{Jozefowicz:2015}, \cite{Lipton:2015}, \cite{Hughes:2013}, and machine translation \cite{Chen:2018}. In parallel, an early form of neural language model was developed in \cite{Bengio:2003}, displaying promising results in statistical language modelling.

LSTM's were first introduced in \cite{Schmidhuber:1997}, specifically to combat the vanishing gradient problem occurring in LSTM model training via Stochastic Gradient Descent. Research has been performed to validate the effectiveness of the LSTM, not only on its ability to ameliorate the vanishing/exploding gradient problem, but also on its ability to capture long term dependencies in text, allowing the model to keep long term explanatory observations to make classifications and predictions in memory. We reference the work of \cite{Krakovna:2016} where they presents promising results in hypothetically combining the HMM and LSTM to increase the predictive accuracy of the models. \cite{Karpathy:2015} showed that both the LSTM and the HMM are capable of identifying special occurrences in text, such as punctuation marks, vowels, etc. In our work, the HMM model is compared with the LSTM model. We compare model performance based on the validation cross-entropy of each LSTM model, and examine the similarities between the HMM and LSTM, in terms of hidden state probabilities.

\subsection{The Hidden Markov Model (HMM)} \label{hmm-section}

Discrete time Hidden Markov Models (HMM) are stochastic models which have a wide range of applications for modelling stochastic processes for various applications. Discrete time HMM’s are ideal for modelling discrete auto-correlated processes, where the observed variables depend on an unobservable hidden state, $S_t$ which obey the Markov property, indicating that the conditional probability of the immediate next state depends only on the present state. The observable symbols at the current time epoch, $t$, are conditionally dependent only on the hidden state at the time $t-1$. HMM’s have been applied in industry to many practical use-cases, most notably in the field of speech recognition \cite{Rabiner:1987}. However, HMM's do extend to a variety of use cases, such as text classification \cite{Krakovna:2016}, fraud detection \cite{Liu:2017}, and autonomous driving \cite{Liu:2019}.

\begin{figure}[h]
\centering
\includegraphics[scale=0.43]{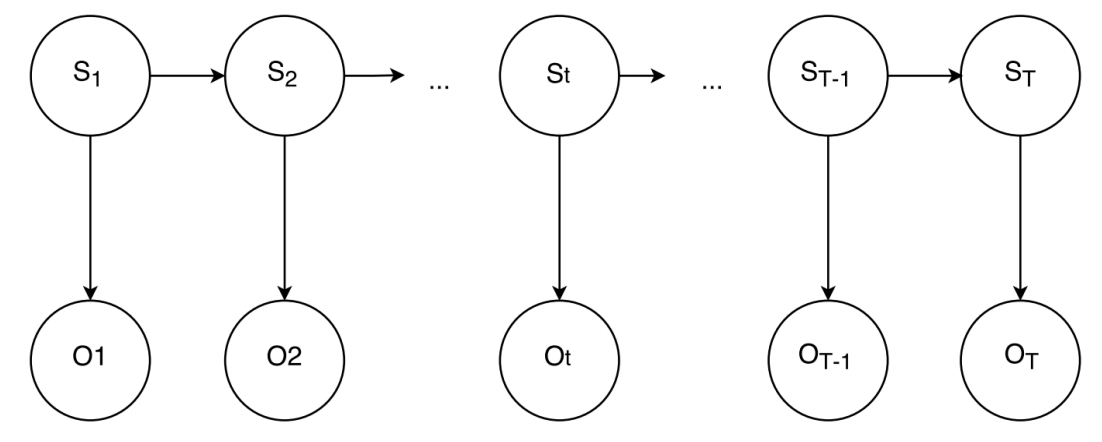}
\caption{Architecture of an HMM \cite{Liu:2017}} \label{hmm-figure}
\end{figure}

We denote the hidden state at time $t$, $S_t$, as a condition representing some underlying state of the sequential text language. The relationship between hidden states, $S_t$, follow a Markov Property, as stated in Eq. (\ref{eq:markov-prop}).

\begin{equation} \label{eq:markov-prop}
	 P(S_{t+1}|S_1,...,S_t) = P(S_{t+1}|S_t) 
\end{equation}

For $M$ observation symbols, and $N$ hidden states, the parameters of the HMM include the transition probability between hidden states denoted as matrix $\pmb{A} = \{a_{ij}\}$. Also the emission matrix $\pmb{B} = \{b_n(m)\}$ for each hidden state, and the initial probability distribution $ \pmb{\pi} = \pi(n_0)$ at time $t = 0$. In order to estimate the parameters of the model we use the Forward-Backward Algorithm outlined in \cite{Rabiner:1987}. The forward probability $\alpha_t(j)$ is defined as the probability of being at state $S_t = j$ while observing sequences $O_{0:t}$, as referenced in Eq. (\ref{eq:forward-alg-2}). And the backward probability, $\beta_t(i)$ is defined as the probability of observing the sequence into the future $O_{t+1:T}$ while currently at state $S_t = i$, as demonstrated in Eq. (\ref{eq:back-alg-2}). The cannonical representation of the forward and backward probabilities are expressed in Equations (\ref{eq:forward-alg-2}) and (\ref{eq:back-alg-2}).

\begin{align}
    \alpha_t({j}) &= b_j(O_t)\sum_{i=1}^N\alpha_{t-1}(i)a_{ij} \label{eq:forward-alg-2} \\
    \beta_t(i) &= \sum_{i=1}^N \beta_{t+1}(j)b_j(O_{t+1})a_{ij} \label{eq:back-alg-2} 
\end{align}

The parameters for multiple sequence HMM's can be estimated using the Baum-Welch Algorithm \cite{Rabiner:1987}. It is possible to estimate the parameters $\pmb{A}$, $\pmb{B}$ and $\pmb{\pi}$, using the Baum-Welch Algorithm for multiple sequences. We define $\gamma_t(i)$ as observing the specific full sequence $O(0:T)$ provided the knowledge of the system being in state $S_t = i$, as defined in Eq. (\ref{eq:gamma}) for $n_h$ hidden states.

\begin{equation}\label{eq:gamma}
    \gamma_t(i) = \frac{\alpha_t({i})\beta_t(i)}{\sum_{i=0}^{n_h}\alpha_t({i})\beta_t(i) }
\end{equation}

We define $\zeta_t(i, j)$ as the probability of $S_t = i$ transitioning to $S_{t+1} = j$ given the sequence of observations $(O_{0:T})$, as defined in Eq. (\ref{eq:zeta}).

\begin{equation}\label{eq:zeta}
    \zeta_t(i, j) = \frac{\alpha_t({i})a_{ij}b_j(S_{t+1})\beta_{t+1}(j)}{P(O_{0:T})}
\end{equation}

Using the Baum-Welch algorithm, iterating through 1 to c iterations, we iterate until convergence for the HMM parameters $\theta_{hmm}^{(c)} = (\pi^{(c)}, A^{(c)}, B^{(c)})$.

\begin{align}
    \pi_i^{(c)} &= \gamma_0(i) \label{eq:initial-param-hmm} \\
    A_{ij}^{(c)} &= \frac{\sum_{t=0}^{T-1}\zeta_t(i,j)}{\sum_{t=0}^{T-1} \gamma_t(i) } \label{eq:transition-param} \\
    B_i(k)^{(c)} &= \frac{\sum_{t=0}^{T} \textbf{1}(O_t = k)\gamma_{t}(i) }{\sum_{t=0}^{T} \gamma_t(i) } \label{eq:emission-mat}
\end{align}

\subsection{Recurrent Neural Networks and Long Short-Term Memory Models (LSTM)} \label{lstm}

Recurrent Neural Networks (RNN) are a class of neural networks capable of processing sequential data using internal memory cells to represent temporal information. Theoretically, RNN's keep track of artificial long-term dependencies; however, back-propagated gradients often vanish or explode as the lengths of sequences grow. Long Short-Term Memory (LSTM), a special architecture of RNN, proposed by \cite{Schmidhuber:1997} was created to alleviate the vanishing gradient problem and retrieve information over long time periods. A LSTM cell consists of three gates: the input, output, and forget gates. The gates serve as regulators that control flow of information within the cell. The core of LSTM follows the equation, expressed in Eq. (\ref{lstm:basic}).

\begin{equation} \label{lstm:basic}
c_t=f_t * c_{t-1}+i_t * g_t
\end{equation}

The cell state, $c_{t}$, serves as the memory of LSTM and enables the RNN to keep track of long-term dependencies. The LSTM cell then chooses to retain a portion of its previous cell state through the forget gate $f_t$, illustrated in Eq. (\ref{lstm:forget}). Another crucial component, the input gate $i_t$, dictates how much new information is incorporated into the updated cell state from the previous cell output, as illustrated in Eq (\ref{lstm:input}). Finally, $g_t$ represents the gate that contains information used to update $c_t$, while $h_t$ serves as the output of the LSTM unit at time $t$, as stipulated in Eq (\ref{lstm:output}). 

\begin{align}
f_t&=\sigma(W_{if}x_t+b_{if}+W_{hf}h_{t-1}+b_{hf}) \label{lstm:forget}\\    
i_t&=\sigma(W_{ii}x_t+b_{ii}+W_{hi}h_{t-1}+b_{hi}) \label{lstm:input}\\  
g_t&=\tanh(W_{ig}x_t+b_{ig}+W_{hg}h_{t-1}+b_{hg}) \\    
h_t&=\sigma(W_{io}x_t+W_{ho}h_{t-1}+b_o)*\tanh(c_{t-1}) \label{lstm:output}
\end{align}

Contrary to the standard RNN architecture, an LSTM cell chooses how much information to retain from previous cells and what to add to the current cell state, $c_t$. As the gradient computations are connected to the forget gate activations, the vanishing gradient problem is greatly alleviated by creating a path for necessary information to flow through $f_t$. The ability to learn additively enables LSTM cells to capture long-term dependencies much more efficiently than standard RNN's do. Since its inception, different flavours of LSTM have been invented to improve upon its existing architectures. Peephole LSTMs \cite{Gerspeep:2000}, which let gate layers borrow information from the cell state, was invented in 2000 to help RNNs learn precise timings. With the rise of computational power in the 2000s, LSTM's achieved record performances in neural language modelling \cite{Graves:2013}. Although LSTM language models perform well, training them on a large dataset is very time-consuming due to normalized distributions of predicted words; forcing the model to consider every word in the dictionary when computing gradients \cite{Jing:2019}. This proven effectiveness and the difficulty in training LSTM language models provide justification into using them as a benchmark for character classification accuracy of this project.

In 2014, a dramatic variation of LSTM, the Gated Recurrent Unit (GRU), was introduced by \cite{cho:2014} to simplify the LSTM architecture. GRU removes the cell state and requires fewer tensor operations to train, which speeds up the training process and provides a viable alternative to LSTM's. An application of this modelling technique could fundamentally speed up our model training period, and could be left to future exploration. Also in recent years, attention-based LSTM's \cite{bahdanau:2014}, which allows for dependencies regardless of the differences in input and output sequences, have achieved impressive model performances in image classification and speaker verification tasks. In particular, the Transformer\cite{vaswani:2017}, a specific type of the attention mechanism, can dramatically speed up the training process and has achieved record-breaking performances in machine translation tasks. Recent developments of LSTM's continue to shed new light on future advancements of RNN's. These advances in RNN architecture could enhance both the modelling accuracy and model training period of our comparative study.

\section{Methodology} 

We compare the performance of the models on the Tiny Shakespeare Dataset (TS) and the War and Peace English Corpus (WP). Each character undergoes one hot encoding of $\nu$ possible alphanumeric characters. The corpus was structured in the form of a sliding window $\pmb{x}$, of $w$ sequential characters, where $n_h$ is the number of hidden states. To be specific, for each window of text, $\pmb{x}$, $\pmb{x}_{0:w-1}$ were selected as the input data, and $\pmb{x}_{1:w}$ as the target variable, therefore providing a one-to-one mapping of a character to the character immediately proceeding it within the sliding window. The recurrent neural networks used in this work were implemented using PyTorch, and the HMM model was implemented using the \textit{hmmlearn} package in Python. The tests were run on a computer with an AMD Ryzen 2600 CPU and an Nvidia GTX 1070 graphics card. 

In this experiment, the HMM model was trained using the Baum-Welch Algorithm. We apply the Viterbi Algorithm to compute the most likely hidden state sequence corresponding to the input observation. Our objective is to compare the vector structure of the hidden state probability distribution with that of the LSTM's counterpart forward pass, on $c_t$. Subsequently, on the same text sequence, we train a corresponding LSTM model with $\nu$ hidden states. 

\section{Experimental Results} \label{sec:result}

The generation of the training data is completed via generating $m$ "batches", \cite{Hinton:2010} refers to this a "mini-batch stochastic gradient descent", and is shown to have convergent behaviour for for convex or pseudo-convex loss functions \cite{Bottou:1998}. We set the mini-batch size $m = 64$ for both training and validation. Sampling is performed to ensure that there is sufficient randomness in the data to make the model training and evaluation robust and fast \cite{Nadeau:2003}.

We measure the performance of the LSTM model based on its ability to accurately predict the subsequent character. We define the $ J(\theta) $ as the cross-entropy loss of the model, as illustrated in Eq. (\ref{eq:cross-ent}). The log-likelihood of a model, $\mathcal{L}(\theta)$, is simply the negative value of the cross-entropy $\mathcal{L}(\theta) = -J(\theta)$.

\begin{equation}
    J(\theta) = - \sum_{i = 1}^{m}\sum_{k = 1}^{\nu} \textbf{1} \{ \pmb{y}_{1:w}^{(i)} = k | \pmb{x}_{0:w-1},  \theta \}  \label{eq:cross-ent}
\end{equation} 

\cite{Krakovna:2016} displayed an improvement to the traditional HMM model and LSTM model in terms of predictive accuracy as measured by predictive log-likelihood of the model, $\mathcal{L}(\theta)$. The LSTM improves its $J(\theta)$ as we increase the complexity of the model by increasing $n_h$, as evidenced in \ref{table:stat_results}. Furthermore, we examined the models seperately and measured the similarity of the hidden states with respect to their cosine distance, $ \Delta(\mathbf{\Psi_H}, \mathbf{\Psi_L}) $. Where $\mathbf{\Psi_H}$ and $\mathbf{\Psi_L}$ corresponds to the probabilities of each of the $n_h$ hidden state vectors.

\begin{equation}
    \Delta(\mathbf{\Psi_H}, \mathbf{\Psi_L} ) = \frac{ \mathbf{\Psi_H} \cdot \mathbf{\Psi_L} }{\big\| \mathbf{\Psi_H} \big\|  \big\| \mathbf{\Psi_L} \big\|  }  \label{eq:cos-dis}
\end{equation} 

In reference to Table \ref{table:stat_results}, it was observed that an increase in the number of hidden states does not guarantee an increase the predictive accuracy of the LSTM. We notice that the cosine similarity, $\Delta(\mathbf{\Psi_H}, \mathbf{\Psi_L} ) $ generally decreases as we decrease the $n_h$. However, between the 15-35 hidden state range for more complex corpora such as WP, there seems to be a subtle increase in the cosine similarity. This indicates that the HMM shares some structural similarities when compared the the LSTM, respective to their hidden state distributions.

    \begin{table}[h!]
    \centering
    \begin{tabular}{||c c c c c||} 
    \hline
    $n_s$ & $\Delta_{TS}$ & $\Delta_{WP}$ & $J(\theta)_{TS}$ & $J(\theta)_{WP}$ \\ [0.5ex] 
    \hline\hline
    5 & 0.792 & 0.741 & 3.724 & 3.950 \\
    10 & 0.801 & 0.657 & 3.476 & 3.831 \\
    15 & 0.816 & 0.613 & 3.581 & 3.678 \\
    25 & 0.715 & 0.674 & 3.246 & 3.497 \\ 
    35 & 0.691 & 0.671 & 3.034 & 3.359 \\
    50 & 0.697 & 0.640 & 3.003 & 3.268 \\
    \hline
    \end{tabular}
    \caption{Effect of the number of Hidden States on $\mathcal{L}(\theta_H)$.}
    \label{table:stat_results}
    \end{table}

Due to the relatively small number of neurons, it was found that training with a GPU was slower than a CPU. This is due to GPU's having a large number of weaker processing units that are well suited for the calculation of high amounts of many small operations, such as linear algebra, that can be highly parallelized \cite{Oh:2004}. Having few neurons in each layer means that there is little parallelization that can be performed. This results in only small improvements to model training speed when using a GPU versus a CPU.


\section{Conclusion}
\label{sec:conclusion}

In this work, we provide a brief survey of the HMM, and LSTM, illustrating the key aspects of stochastic modelling and deep learning for the purpose of language modelling and text prediction. We further demonstrate the effectiveness of each model seperately, and demonstrate the similarities of the hidden state structure of the HMM and LSTM respectively. We observe that there are structural similarity between the hidden states of the LSTM and HMM, measuring the hidden state probability output vectors of both models. From the measurement, we find that general similarities do exist, especially at lower dimensionalities. However, in complex corpora such as the War and Peace dataset, the cosine similarity may increase with an increase in hidden state dimensionality. The results of this paper justify that the HMM can be used as a suitable approximator for the LSTM, and using this similarity one can develop a hybrid model which may simplify the model training behaviour of the LSTM or increase its respective performance. Furthermore, an area of improvement can to modify our GPU based code in order to improve the training times of the model, and not rely on CPU based computation. In summary, this work has advanced the understanding of stochastic and deep learning based models aimed at classifying text patterns.  

\bigskip
\bibliographystyle{aaai}\bibliography{example.bib}
\end{document}